\documentclass[11pt]{article}

\usepackage[preprint]{acl}

\usepackage{times}
\usepackage{latexsym}

\usepackage[T1]{fontenc}

\usepackage[utf8]{inputenc}

\usepackage{microtype}

\usepackage{graphicx}
\usepackage{booktabs}
\usepackage{amsmath}
\usepackage{amssymb}
\usepackage{xcolor}
\title{Ill-Posed by Design: Probing Evidence Use in VLMs}

\author{
  \textbf{Boaz Meivar\textsuperscript{*,1}},
  \textbf{Shaked Perek\textsuperscript{*,2}},
  \textbf{Shani Shvartzman\textsuperscript{*,1}},
  \textbf{Eli Schwartz\textsuperscript{2}},
  \textbf{Shai Avidan\textsuperscript{1}}
\\
\\
  \textsuperscript{1}Tel Aviv University,
  \textsuperscript{2}IBM Research
\\
  \small{
    \textbf{Correspondence:} \href{mailto:boazmeivar@mail.tau.ac.il}{boazmeivar@mail.tau.ac.il}
  }
\\
  \small{\textsuperscript{*}Equal contribution.}
}

\begin{document}
\maketitle
\begin{abstract}
Counterfactual analysis is widely used to study evidence use in vision-language models, but its diagnostic value is limited on well-posed tasks: when several cues independently support the same answer, removing one may not change the prediction. We propose monocular metric object-size estimation as an ill-posed diagnostic setting for evidence selection: because physical size cannot be determined from a single uncalibrated image, models must rely on imperfect cues---category priors, target appearance, local context, apparent image size, and scene geometry. We assemble \textbf{Metric VQA} ($10{,}813$ dimension queries from Objectron and $331$ tape-measured in-the-wild scenes) and evaluate $12$ open-weight VLMs ($3$--$397$\,B parameters) with counterfactual analysis decomposing six visual and language evidence channels. Even the largest VLMs tested (Qwen3-VL-235B, Qwen3.5-397B, InternVL3.5-241B) trail a \textbf{text-only} frontier LLM on the in-the-wild split. The diagnostic analysis shows: target identity is the most load-bearing cue, target pixels and local context help only some models, apparent size shifts predictions without a directional readout, and global scene geometry is largely unused. We analyze LoRA fine-tuning as an actionable intervention specific to metric estimation: while the task is learnable, the models do not learn to leverage scene geometry.
\end{abstract}

\begin{figure*}[t]
\centering
\includegraphics[width=\textwidth]{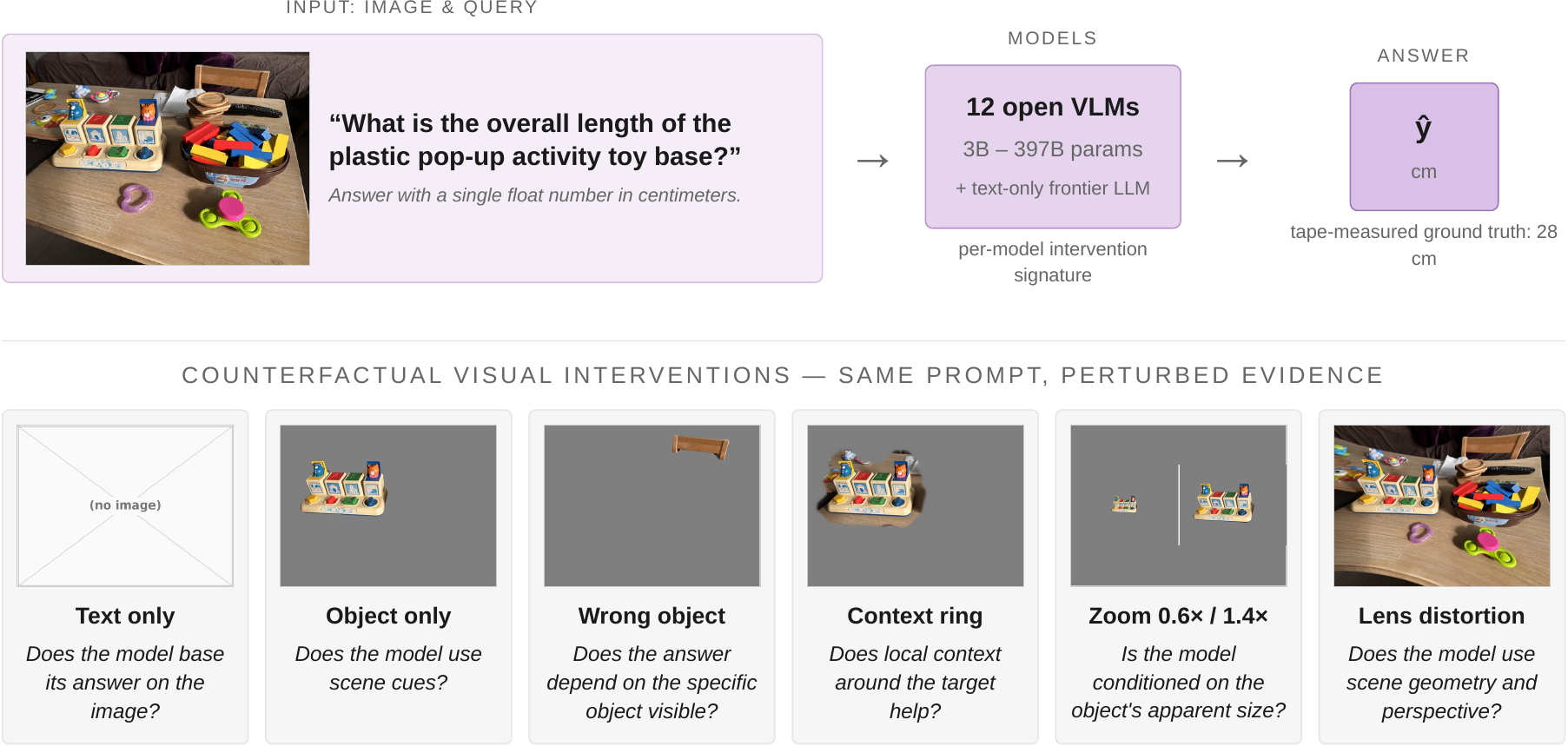}
\caption{\textbf{Ill-posed metric estimation as a diagnostic for evidence use in VLMs.} We apply six image-level interventions (object-only mask, wrong-object substitution, context ring, zoom $0.6\times$/$1.4\times$, lens distortion) and a text-only ablation, holding the query fixed and re-querying $12$ open-weight VLMs spanning $3$--$397$\,B parameters. Each panel pairs the intervention with the headline finding across the pool; full per-model signatures are in Section~\ref{sec:results}.}
\label{fig:teaser}
\end{figure*}

\section{Introduction}

Counterfactual interventions are a routine tool for analyzing how machine-vision models use evidence: perturbing or masking specific image regions and reading the resulting change in model output is widely used to attribute predictions to candidate cues \citep{fong2017perturbation, zhao2022vlchecklist}.

However, the diagnostic power of such interventions is limited on well-posed visual tasks. For example, in object-presence recognition, multiple visual cues can independently justify the same label. Masking a single cue may therefore leave the prediction unchanged, making it difficult to determine whether the masked cue was unused or merely redundant with other evidence.

This paper introduces ill-posed tasks as a more informative diagnostic setting: because no single cue fully determines the answer, controlled perturbations better expose which evidence a model relies on. A prediction that shifts under intervention indicates the cue was load-bearing.

We instantiate this setting with monocular metric object-size estimation. The answer is underdetermined by a single image: without calibration, stereo, or known reference objects, physical scale is not directly observable, so models must rely on imperfect cues such as category-level size priors, nearby objects, apparent image size, or perspective. Standard accuracy metrics collapse these strategies into a single score: a correct answer may come from a prior rather than image understanding, and an error does not reveal which cue failed.

We assemble \textbf{Metric VQA}, a benchmark of $10{,}813$ dimension queries on two splits: $10{,}482$ queries on Objectron frames with 3D-bounding-box ground truth, and $331$ tape-measured queries on cluttered in-the-wild scenes. We evaluate $12$ open-weight VLMs ($3$--$397$\,B parameters) against a frontier text-only LLM and a human reference, applying counterfactual analysis across six evidence channels: category priors, target pixels, target identity, local context, apparent image size, and global scene geometry.

Even the largest open VLMs we tested (Qwen3-VL-235B, InternVL3.5-241B, Qwen3.5-397B) trail a text-only frontier LLM with no image access on the in-the-wild split of Metric VQA. The intervention signatures reveal a stratified pattern: target identity is load-bearing, target pixels and local context are leveraged by only a subset of models, apparent size shifts predictions without a directional readout, and global scene geometry is largely unused. We further analyze supervised LoRA fine-tuning as an actionable intervention specific to metric estimation: the task is learnable, but gains do not reflect new visual-geometry understanding. Instead, fine-tuning improves the channels each base model already partially uses (target identity, local context, language-prior access), with the specific improvement differing by model.

\paragraph{Contributions.}
\begin{itemize}
\item We frame monocular metric object-size estimation as an ill-posed diagnostic for evidence selection in VLMs, and introduce a suite of counterfactual visual interventions probing six evidence channels: category priors, target pixels, target identity, local context, apparent image size, and global scene geometry.
\item A text-only frontier LLM with no image access outperforms every open-weight VLM we tested ($3$--$397$\,B parameters) on the in-the-wild split of \textbf{Metric VQA}.
\item Intervention signatures show that target identity is the most consistently load-bearing visual cue; target pixels and local context are leveraged by only a subset of models; apparent size affects predictions but not in a direction-consistent way; and global scene geometry is largely unused.
\item We analyze LoRA fine-tuning as an actionable intervention specific to metric estimation, showing that while this ill-posed task is learnable, the models do not learn to leverage scene geometry.
\item We release \textbf{Metric VQA} (data, intervention variants, predictions, code) to support reproducible diagnostic analysis of evidence use in multimodal systems.
\end{itemize}

\section{Related Work}
\label{sec:related}

\paragraph{Quantitative reasoning in VLMs.}
Single-image metric inference has long been recognized as ill-posed without calibration or explicit reference \citep{eigen2014depth}, yet recent work has increasingly pushed VLMs toward this regime.
SpatialVLM \citep{chen2024spatialvlm} curates internet-scale 3D-grounded VQA data and co-trains a VLM to improve qualitative and quantitative spatial reasoning, including distances and size differences.
DepthLM \citep{depthlm} trains VLMs for metric depth prediction using explicit scale supervision and camera normalization, showing that strong metric depth performance can be obtained with sparse text-style supervision and visual prompting.
These works are capability-building efforts: they improve metric reasoning through additional training or supervision.
Our work is diagnostic rather than methodological: rather than training VLMs for metric tasks, we apply targeted counterfactual interventions to characterize what evidence current models use under monocular ambiguity.

The closest existing benchmark on single-image metric estimation is Q-Spatial Bench \citep{liao2024qspatial}, which evaluates VLMs on quantitative size and distance questions and introduces SpatialPrompt, a zero-shot prompting method that encourages reference-object reasoning.
Their intervention is at the \emph{prompt} level: they elicit a reasoning pattern and show that it improves performance.
Our work is methodologically orthogonal: we intervene at the \emph{image} level, manipulating category prior, target pixels, target identity, apparent size, local context, and global scene geometry while keeping the prompt fixed.
QuantiPhy \citep{quantiphy} is closest in diagnostic spirit: it studies video-based estimation of kinematic quantities such as size, velocity, and acceleration, and finds that VLMs often rely more on pretrained priors than on visual measurements.
We complement this direction with a single-image, object-centric setting in which scale is fundamentally ambiguous and visual evidence can be decomposed through targeted counterfactual interventions.

\paragraph{Language priors and blind baselines.}
A recurring lesson in multimodal evaluation is that strong VLM scores can hide reliance on non-visual signals.
\citet{goyal2017vqa} introduced VQA v2 using complementary images that share a question, showing that earlier models exploited answer-side language priors rather than visual evidence; \citet{agrawal2018gvqa} extended this idea to changed-prior splits and architectural debiasing.
More recently, SugarCrepe \citep{hsieh2023sugarcrepe} showed that \emph{blind} text-only models can outperform state-of-the-art VLMs on widely used compositional benchmarks, exposing benchmark shortcuts.
We extend this blind-baseline perspective to quantitative physical estimation: object-size questions appear intrinsically visual, yet strong priors alone can explain a substantial part of model performance.

\paragraph{Diagnostic and counterfactual multimodal evaluation.}
A separate line of work decomposes VLM behavior through controlled probes rather than aggregate scores.
VL-CheckList \citep{zhao2022vlchecklist} evaluates pretrained VLMs along independently manipulable axes such as objects, attributes, and relations.
Causal saliency methods such as \citet{fong2017perturbation} use input perturbations and the resulting output changes as a basis for explanation.
We adapt this perturbation logic to multimodal quantitative reasoning, defining interventions over semantic and geometric evidence channels---target pixels, target identity, apparent size, local context, and scene geometry---rather than pixel-level relevance maps.
The result is a diagnostic instrument for evidence selection, not a benchmark for ranking.

\section{Benchmark and Setup}
\label{sec:setup}

We formalize metric object-size estimation as visual question answering: given a single RGB image and a natural-language query about a physical dimension (e.g., \emph{``How tall is this microwave?''}), the model must produce a numeric answer in centimeters. We assemble \textbf{Metric VQA}, a benchmark of $10{,}813$ dimension queries paired with ground-truth measurements, partitioned into two complementary splits: an Objectron-derived curated split of $10{,}482$ queries spanning $9$ everyday object categories with 3D-bounding-box ground truth, and an in-the-wild split of $331$ queries on cluttered scenes containing multiple objects, with tape-measured ground truth.

\paragraph{Task formulation.}
Each evaluation instance is a triple $(\mathbf{I}, q, y)$, where $\mathbf{I}$ is the RGB image, $q$ is the dimension query, and $y \in \mathbb{R}^{+}$ is the ground-truth dimension in centimeters. The model receives $\mathbf{I}$ together with $q$ followed by the instruction \emph{``Answer with a single float number in centimeters''}. A numeric prediction $\hat{y}$ is extracted by a deterministic regex parser that returns the first float or integer token in the response, following the relaxed-exact-match convention of~\citet{masry2022chartqa}. Responses with no parsable number are recorded as \emph{parse failures}; for aggregate MAE we additionally clip predictions above $10^{4}$\,cm to suppress occasional numerical blowups (e.g.\ scientific notation artifacts) that would otherwise dominate the mean.

\subsection{Objectron-derived split}
\label{sec:objectron}
We derive a large curated split from Google Objectron~\citep{ahmadyan2021objectron}, which provides object-centric video frames with camera poses and oriented 3D bounding boxes across nine everyday categories: \emph{bike, book, bottle, camera, cereal box, chair, cup, laptop, shoe}. From each annotated frame we extract the three physical dimensions (height, width, depth) of the 3D bounding box, yielding one dimension query per axis per frame. After quality filtering, the final split comprises $3{,}494$ frames and $10{,}482$ dimension queries, with $27$ unique (category, axis) query templates.

\subsection{In-the-wild cluttered scenes}
\label{sec:itw}
To evaluate performance in a less constrained setting, we collected $70$ scene photographs spanning everyday environments (kitchens, living rooms, desks, outdoors). Unlike the object-centric Objectron frames, in-the-wild (ITW) images contain multiple objects with arbitrary occlusion, lighting, and viewpoint. For each image we formulated between $1$ and $8$ dimension queries targeting objects from $0.8$ to $245$\,cm, with every ground-truth dimension recorded by tape measure.

The resulting split contains $331$ queries on $70$ scenes covering $\sim$$99$ distinct objects (Figure~\ref{fig:dataset_composition}, Appendix~\ref{sec:appendix}), spanning a wider range of physical sizes than Objectron's $9$ curated categories. This is comparable in scope to the closest peer-reviewed manually-measured benchmark, the Q-Spatial++ split of Q-Spatial Bench ($101$ questions on $87$ scenes; \citealp{liao2024qspatial}).

As a reference we provide human-annotated performance: a single annotator (not involved in data collection) estimated all $331$ dimensions from the images alone.

\subsection{Evaluation metrics}
\label{sec:metrics}
We report \textbf{Acc@10\%} as our primary metric: the fraction of queries with $|\hat{y}-y|/y \leq 0.1$. It is invariant to object scale and rewards calibrated predictions. MAE (mean absolute error in centimeters) is reported alongside in Appendix~\ref{sec:appendix}, Table~\ref{tab:main_mae}.

For paired intervention analyses we additionally compute a \emph{paired prediction ratio} $\hat{y}_{\text{in}}/\hat{y}_{\text{out}}$ between zoom-in and zoom-out variants of the same object. A scale-invariant model yields a ratio near $1.0$. Where an intervention produces a small $\Delta$ Acc@10\% relative to a paired anchor (most notably the lens-distortion probe), we report a $95$\% paired-bootstrap confidence interval ($B{=}2000$ resamples of paired-row indices) on the per-model $\Delta$, and use the CI rather than a fixed threshold to classify each model.

\subsection{Models and baselines}
\label{sec:models}
We evaluate \textbf{12 open-weight VLMs} spanning roughly two orders of magnitude in scale ($3$\,B to $397$\,B parameters): qwen family models: Qwen2.5-VL-3B, Qwen2.5-VL-7B~\citep{bai2025qwen25vl}, Qwen3-VL-235B-A22B-Instruct ($235$\,B mixture-of-experts) and Qwen3.5-397B ($397$\,B); InternVL2-8B and InternVL3.5-241B~\citep{chen2024internvl}; LLaVA-v1.6-Mistral-7B and LLaVA-OneVision-7B~\citep{liu2023llava}; Molmo-7B-D~\citep{deitke2025molmo}; SmolVLM~\citep{marafioti2025smolvlm}; Gemma3-4B, and granite-vision-3.3-2B. As an image-free language baseline we query a frontier LLM (\textbf{ChatGPT~5.5}~\citep{openai2024chatgpt}) with the dimension question alone and no image; this measures what is achievable from the natural-language category prior. On ITW split we additionally report the human baseline described in Section~\ref{sec:itw}.

\section{Counterfactual Analysis}
\label{sec:interventions}

Each intervention takes an (image, query) pair from ITW split and produces a modified image while leaving the natural-language query unchanged. The same VLM is then queried with the modified image. Pixels that we wish to ``remove'' are replaced with a uniform gray fill at intensity $127$. Per-object target masks are obtained with SAM 3~\citep{carion2025sam3} from each query's noun phrase.

\subsection{Language-only priors}
\label{sec:int_blank}
The cleanest probe of a VLM's internal language prior is to remove the image modality from the prompt entirely (\emph{text-only}). Molmo-7B-D refuses text-only (returns empty on every query). In this case we input a uniform gray image (RGB $127$) as a text-only proxy, providing no visual cue on which the model may ground its answer. Text-only input is reported for all other models. As an external reference we additionally query a frontier text-only LLM (ChatGPT~5.5) with the dimension question and no image at all, measuring what a strong language-only prior can extract from the query alone.

\subsection{Target pixels, identity, and local context}
\label{sec:int_context}
Four interventions share the same SAM-mask backbone and differ in which mask is kept visible: \textbf{object\_only} fills gray outside the target's mask, leaving only target pixels; \textbf{object\_ring} fills gray outside a $50$\,px dilation of the mask, keeping a thin ring of local context around the object; \textbf{wrong\_object} fills gray outside a \emph{different} queried object's mask on the same image, swapping target identity while preserving location and scene statistics; \textbf{random\_mask} fills gray outside a random rectangle whose pixel area matches the target mask, controlling for the amount of visible pixels. Together these four interventions separate the contributions of target pixels (\textsc{object\_only} vs.\ \textsc{random\_mask}), target identity (\textsc{wrong\_object} vs.\ \textsc{object\_only}), and local context (\textsc{object\_ring} vs.\ \textsc{object\_only}).

\subsection{Apparent size}
\label{sec:int_zoom}
To test whether predictions track the object's pixel footprint, we rescale the object-only image by $1.4\times$ (zoom-in) or $0.6\times$ (zoom-out) and crop or pad back to the original canvas on the mask centroid. The two factors give a $1.4/0.6 \approx 2.33\times$ change in the target's apparent linear scale, so a model whose predictions track image-space linear size should produce a paired ratio $\hat{y}_{1.4\times}/\hat{y}_{0.6\times}$ near $2.33$; a scale-invariant model produces a ratio near $1.0$.

\subsection{Radial lens distortion}
\label{sec:int_distortion}
To probe sensitivity to global scene geometry, we warp each full ITW image with a radial distortion model $r' = r(1 + k r^2)$, where $r$ is the normalized distance from image center. Positive $k$ gives barrel distortion (curves straight lines outward), negative $k$ pincushion (inward); both perturb vanishing-point structure while preserving object identity and approximate apparent size. We report both directions at $|k| = 0.15$.

\section{Analyzing an Actionable Intervention for Metric Estimation}
\label{sec:lora_intervention}

Supervised LoRA fine-tuning is the actionable, metric-estimation-specific intervention we analyze in this paper. To diagnose what it changes, we re-apply the counterfactual suite of Section~\ref{sec:interventions} to the LoRA-fine-tuned models. This section describes the fine-tuning setup; the base-vs-finetuned comparison is reported in Section~\ref{sec:lora}.

We fine-tune on the Objectron split, holding out $500$ randomly sampled queries for validation and selecting the best checkpoint by validation loss. ITW split is held out entirely; any LoRA improvement on it therefore measures transfer from curated single-object frames to cluttered multi-object scenes. We attach LoRA~\citep{hu2021lora} adapters (rank $r{=}8$, $\alpha{=}16$, dropout $0.05$) to all linear projections in both the vision encoder and the language model, covering attention (Q/K/V/O) and MLP/FFN layers; the vision encoder is not frozen, since the intervention suite probes how each model uses image-side evidence. We train for $2$ epochs at constant lr $=1\mathrm{e}{-}5$, batch size $1$ with gradient accumulation $4$, standard next-token cross-entropy on the answer tokens, and fp16 precision. Each model is trained on a single H100 80GB GPU. Training takes approximately 15 minutes for 3B models and 2–3 hours for 7B models (~9 GPU-hours total across all models). All VLM inference and LoRA training use HuggingFace Transformers; LoRA adapters are implemented with the PEFT library.

\section{Results and Discussion}
\label{sec:results}

We structure the results around the evidence sources probed by our
interventions, not as a leaderboard. Table~\ref{tab:main} reports
Acc@10\% on both splits: large VLMs exceed the language-prior baseline
(ChatGPT~5.5 text-only) on Objectron but not on the ITW split,
with consistent split rankings (Spearman $\rho = 0.94$, $p < 0.001$).
These scores establish the central diagnostic problem---a correct
estimate may come from category knowledge alone, and an incorrect one
does not reveal which channel failed (target recognition, local
context, apparent size, or scene geometry). We therefore analyze each
counterfactual intervention from Section~\ref{sec:interventions} as a
paired test of evidence use. Per-model decomposition into parse
failures, scale failures, and committed estimations is in
Appendix~\ref{sec:appendix}, Table~\ref{tab:fse_itw}.

\begin{table}[t]
\centering\small
\begin{tabular}{l|cc}
\toprule
Model & ITW Acc@10\% & Obj Acc@10\% \\
 & ($n{=}331$) & ($n{=}10{,}482$) \\
\midrule
\multicolumn{3}{l}{\emph{Non-VLM reference baselines}} \\
Human reference & 39.3\% & -- \\
ChatGPT 5.5 (text) & 36.9\% & 32.4\% \\
\midrule
Qwen3-VL-235B & 30.8\% & 38.9\% \\
Qwen3.5-397B & 30.8\% & 43.6\% \\
InternVL3.5-241B & 20.9\% & 30.3\% \\
Qwen2.5-VL-7B & 20.8\% & 23.9\% \\
Molmo-7B-D & 19.3\% & 21.4\% \\
InternVL2 & 16.9\% & 24.8\% \\
LLaVA-OV & 16.6\% & 20.8\% \\
LLaVA-v1.6 & 13.6\% & 18.1\% \\
Gemma3-4B & 12.7\% & 18.5\% \\
granite-3.3-2b & 11.4\% & 9.0\% \\
SmolVLM & 7.3\% & 12.0\% \\
Qwen2.5-VL-3B & 6.0\% & 9.8\% \\
\bottomrule
\end{tabular}
\caption{Acc@10\% on both splits, sorted by in-the-wild. Non-VLM rows are reference baselines, not benchmark contenders. MAE counterpart in Appendix~\ref{sec:appendix}, Table~\ref{tab:main_mae}.}
\label{tab:main}
\end{table}

\begin{figure*}[t]
\centering
\includegraphics[width=\textwidth]{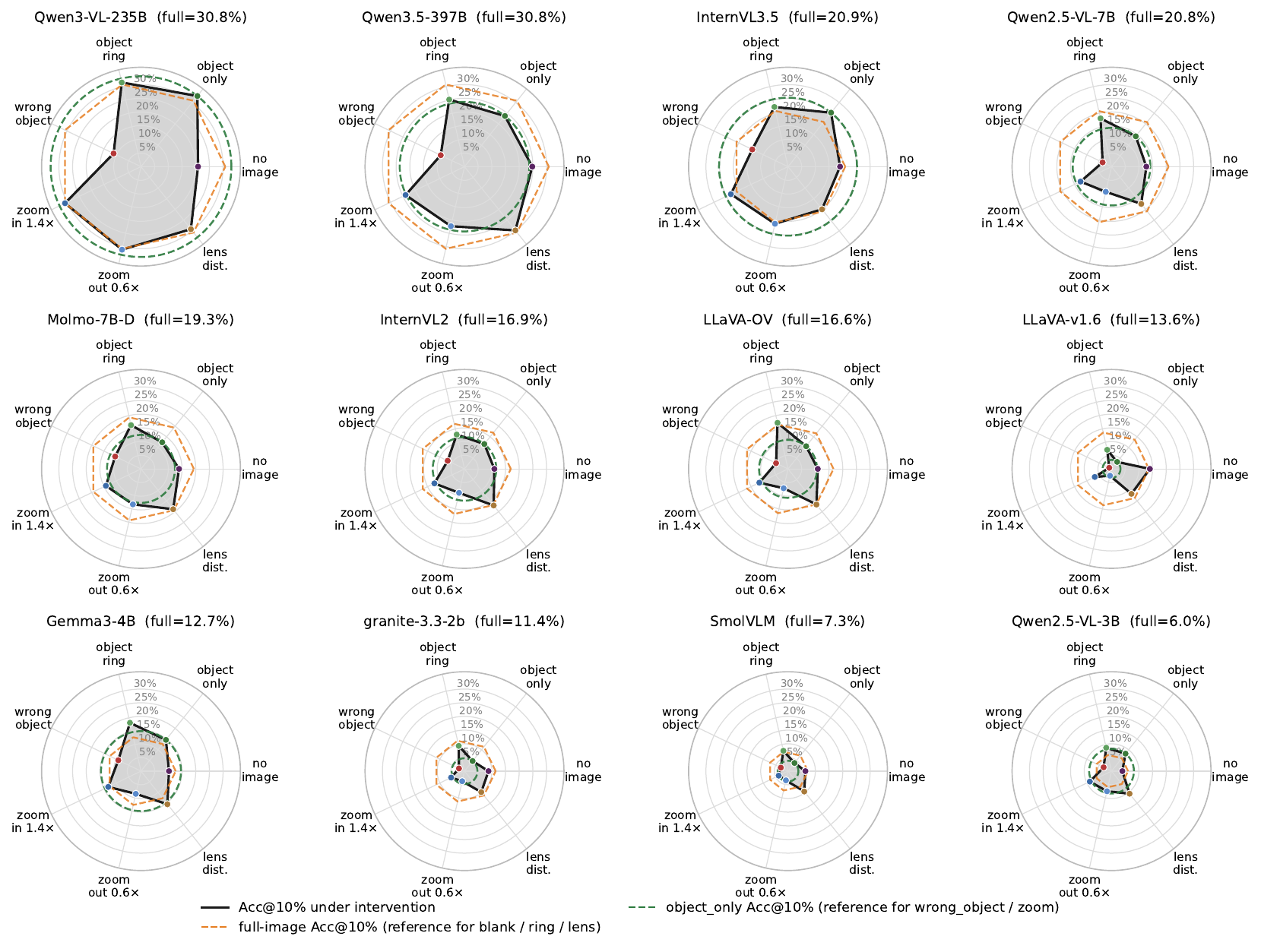}
\caption{\textbf{Per-VLM intervention signature on the in-the-wild split.} Each panel is one VLM; axes are the interventions defined in Section~\ref{sec:interventions}. The solid black polygon traces Acc@10\% on each intervention. The \textcolor{orange!85!black}{orange dashed hexagon} marks the model's full-image Acc@10\% (anchor for no-image / object\_ring / lens dist.); the \textcolor{green!50!black}{green dashed circle} marks its object\_only Acc@10\% (anchor for wrong\_object / zoom-in / zoom-out). A vertex inside the relevant anchor indicates degradation under that intervention.}
\label{fig:intervention_radar}
\end{figure*}

\subsection{Language priors persist without images}
\label{sec:res_prior}

The language-prior probes ask how much of each model's prediction can be explained without visual evidence. Text-only is the primary probe for every model except Molmo-7B-D, which refuses text-only and is reported on gray-blank. The strongest open-VLM language-prior measurement (Qwen3.5-397B text-only) still falls below ChatGPT~5.5 on both splits.

The key result is that no open VLM \emph{with} the image beats the frontier LLM \emph{without} the image on ITW split, and only the two largest open VLMs exceed it on Objectron. This is diagnostically important: monocular metric estimation sounds intrinsically visual, yet a strong language prior explains much of the achievable performance. The intervention analyses below ask when, and for which models, the image adds evidence beyond that prior. Per-model numbers and the image-vs-prior scatter are in Appendix~\ref{sec:appendix} (Table~\ref{tab:language_prior}, Figure~\ref{fig:prior_vs_pixel}).

\subsection{Target evidence: identity is the strongest visual cue}
\label{sec:res_context}

We evaluate the mask-based interventions using the paired comparisons defined in Section~\ref{sec:int_context}: \textsc{object\_only} vs.\ \textsc{random\_mask} for target pixels, \textsc{wrong\_object} vs.\ \textsc{object\_only} for target identity, and \textsc{object\_ring} vs.\ \textsc{object\_only} for local context. Note that \textsc{object\_only} also has a single-counterfactual reading on its own (the scene is removed, the target preserved); we report both perspectives below. The full per-model table is in Appendix~\ref{sec:appendix}, Table~\ref{tab:context}.

Among tested models, target identity collapse is the most reliable image-side effect: replacing the queried object with a different object produces an accuracy drop whose $95\%$ CI lies below zero for $8$ of $12$ models, with the largest drops in the larger models (largest: Qwen3-VL-235B at $-20.3$\,pp). This indicates that many VLMs are not merely answering from the query text: they condition their estimate on which object is visible.

The stronger half of the tested models leverages information from target pixels for metric estimation. The \textsc{object\_only} condition significantly improves over \textsc{random\_mask} for $6$ of $12$ models, all from the larger pool (largest lift: Qwen3-VL-235B at $+15.7$\,pp). For the remaining $6$ models, the target crop is not statistically more useful than a random patch of matched area, suggesting weak extraction of metric information from target appearance.

Local context is leveraged by a mid-tier subset of the tested models: adding a $50$\,px halo around the object lifts Acc@10\% with a CI strictly above zero for $5$ of $12$ models---all with weaker pixel-level use (largest lift: LLaVA-OV at $+6.7$\,pp).

Together, these results give a stratified evidence map: target identity is the most reliably load-bearing channel, target pixels help mainly for stronger models, and local context provides a smaller, model-dependent benefit. Notably, \textsc{object\_only} matches or exceeds the full-scene Acc@10\% for several models---Qwen3-VL-235B ($33.0\%$ vs $30.7\%$), InternVL3.5-241B, Gemma3-4B---suggesting that on ITW split, scene clutter is at least as often a distractor as an informative cue.

\subsection{Apparent size: VLMs do not read off image-space scale}
\label{sec:res_zoom}

We find that VLM predictions change with apparent size, but not in the direction of the shift---larger apparent size does not systematically push predictions upward, and smaller apparent size does not systematically push them downward. We rescale the object-only image by $1.4\times$ (zoom-in) and $0.6\times$ (zoom-out); a direct readout of apparent size would give a paired prediction ratio near $2.33$, but no model in our pool shows this. Under zoom-in, models are robustly invariant: median paired ratio is $1.00$ for all $12$ models, and Acc@10\% deltas relative to \textsc{object\_only} stay within $\pm 2$\,pp. Under zoom-out (Table~\ref{tab:apparent_size_bias}, Appendix~\ref{sec:appendix}), the cue destabilizes predictions for nearly half the pool: five models lose $3$--$6$\,pp Acc@10\% versus \textsc{object\_only} (Gemma3-4B $-6.1$\,pp, Qwen2.5-VL-7B $-4.8$\,pp, InternVL3.5-241B $-3.8$\,pp, LLaVA-OV $-3.4$\,pp, InternVL2 $-2.6$\,pp)---predictions are clearly changing under the intervention. Yet only Gemma3-4B ($r{=}0.80$) and LLaVA-OV ($r{=}0.83$) shrink predictions in the cue's direction; for the other models the shifts are non-directional. Apparent size therefore acts as a destabilizing signal for several models rather than as a metric cue.

\subsection{Scene geometry: weak perspective use}
\label{sec:res_distortion}

Scene geometry is the weakest and least consistent evidence channel in our suite: only $1$ of $12$ models shows a reliable Acc@10\% drop under radial lens distortion, and the sign of the shift does not pattern with the direction of distortion for any model. Concretely, only Qwen2.5-VL-7B has a $95\%$ CI strictly below zero under both barrel and pincushion distortion ($-3.5$\,pp and $-5.3$\,pp); for eight models both CIs cover zero, and three are mixed (CI excludes zero in one direction, includes zero in the other). A VLM that used vanishing-point structure to infer scale would, in principle, degrade more in one direction than the other---we see no such pattern. Per-model $\Delta$\,Acc@10\% with paired-bootstrap CIs is in Appendix~\ref{sec:appendix}, Table~\ref{tab:distortion}.

\subsection{What LoRA changes (and what it doesn't)}
\label{sec:lora}

Our finding show that supervised LoRA fine-tuning improves accuracy on metric estimation across all three VLMs we test, but the gains do not reflect new geometric understanding of the scene. Instead, fine-tuning improves channels each base model already partially uses (target identity, local context, language-prior access), with the specific improvement differing by model.

\begin{figure*}[t]
\centering
\includegraphics[width=\textwidth]{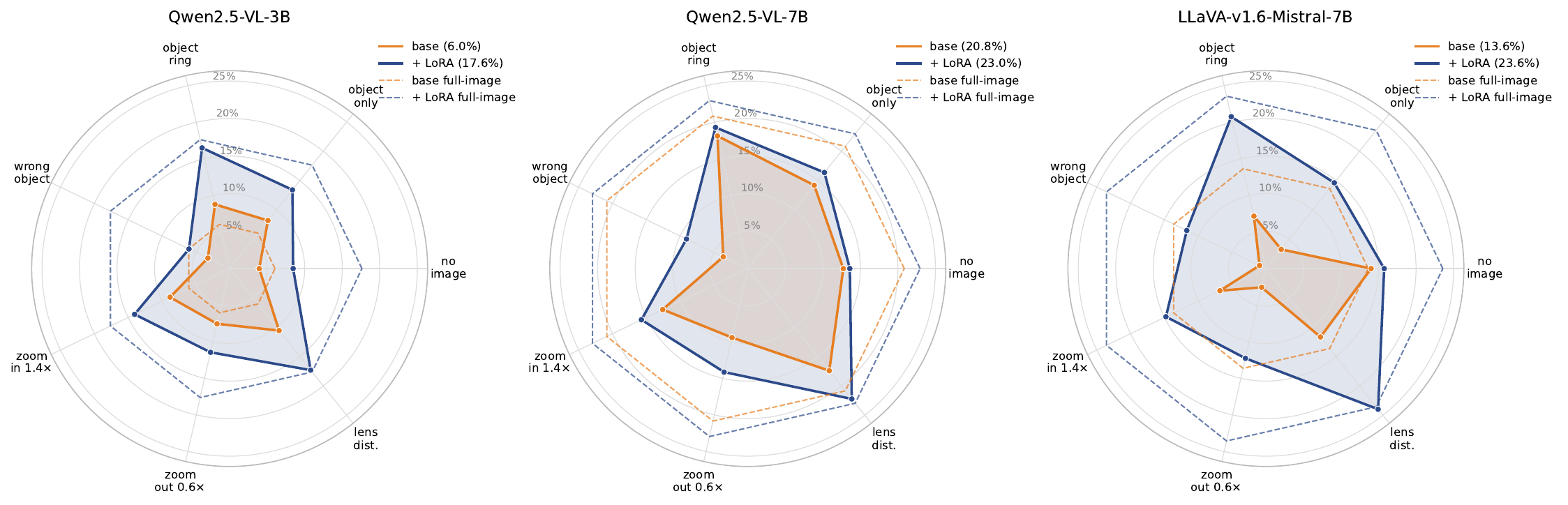}
\caption{\textbf{Per-model base-vs-LoRA intervention signatures.} Each panel overlays base (orange) and LoRA-finetuned (blue) Acc@10\% on the intervention axes of Figure~\ref{fig:intervention_radar}. Dashed contours mark each model's full-image Acc@10\% as a within-model reference (orange = base, blue = LoRA).}
\label{fig:lora_radar}
\end{figure*}

We train LoRA adapters on three open VLMs spanning the lower-to-mid range---Qwen2.5-VL-3B (base $6.0\%$), LLaVA-v1.6-Mistral-7B ($13.6\%$), and Qwen2.5-VL-7B ($20.8\%$)---and re-run the full intervention suite (Figure~\ref{fig:lora_radar}). All three lift on full-image Acc@10\% ($+11.6$, $+9.9$, $+2.1$\,pp), but the per-channel decomposition reveals three distinct mechanisms. \textbf{Qwen2.5-VL-7B}, already image-engaged at base, mainly \emph{stabilizes the language prior}: every impoverished-image condition lifts to within $2$--$5$\,pp of the model's text-only Acc. \textbf{Qwen2.5-VL-3B}'s base treats the full scene as a distractor (\textit{object\_only} $>$ full); LoRA \emph{flips this relation} and induces target-identity sensitivity (the wrong-object penalty deepens). \textbf{LLaVA-v1.6}---the only model where the base image is net-harmful (text-only $14.0\%$ $>$ vision $13.6\%$)---has LoRA \emph{introduce target-identity and local-context use} (\textit{object\_only} $+11.4$\,pp; \textit{object\_ring} $+13.8$\,pp), shifting the image from $-0.4$ to $+7.9$\,pp of value. Across all three, LoRA does not introduce new apparent-size sensitivity or perspective-cue use that the base lacked.

\subsection{Memory over visual evidence}
\label{sec:res_synthesis}

Read together, the interventions show that VLM object-size estimates are neither pure language priors nor reliable visual measurements. Models often use the image, especially target identity, but visual evidence is selective and inconsistent: target pixels and local context help only for subsets of models, apparent size shifts predictions without a directional readout, and global scene geometry has little consistent effect. Light LoRA fine-tuning learns the task on the models we tested but patches existing weak channels rather than introducing new visual-geometry use.

A VLM's centimeter estimate should not be treated as a measurement. It is better understood as a category-level prior with selective, model-dependent visual modulation. This is precisely what makes monocular metric estimation diagnostic: underdetermined scale forces models to choose among imperfect evidence sources, and intervention signatures reveal those choices more clearly than aggregate accuracy.

\section{Conclusion}
\label{sec:conclusion}

We frame monocular object-size estimation as an ill-posed diagnostic for evidence selection in VLMs. Across $12$ open-weight VLMs, intervention signatures reveal a stratified pattern: target identity is the most consistently load-bearing visual cue, while apparent size shifts predictions without a directional readout and global scene geometry is largely unused. Even the largest open VLMs we tested trail a text-only frontier LLM with no image access on the in-the-wild split of Metric VQA. LoRA fine-tuning makes the task learnable but improves existing weak channels rather than introducing new visual-geometry use.

\paragraph{Takeaways.} Leveraging an ill-posed task as a diagnostic tool is highly informative in multimodal machine vision: when cues are not redundant, counterfactual interventions isolate cue use rather than test for redundancy, exposing patterns that aggregate accuracy hides. Research progress is often measured by leaderboard performance, yet our LoRA analysis reveals significant shortcut behavior---accuracy improves through cues the model already partially uses, not through new visual-geometry capability. Asking \emph{why} a model improved is at least as informative as reporting that it did.

\paragraph{Future work.} A natural next step is to test whether the missing visual-geometry channel can be induced by training that directly rewards its use---e.g., supervision that penalizes invariance to scene-geometry perturbations, or reinforcement signals tied to counterfactual consistency. Our LoRA results show that next-token-loss fine-tuning alone does not achieve this; whether a more targeted signal can shift VLMs from prior-conditioned readouts to genuine scene-geometry use is an open question. The same ill-posed diagnostic framework also extends to other underdetermined visual tasks---depth, distance, mass, count, material---where individually load-bearing cues invite the same per-channel decomposition.

We release \textbf{Metric VQA}---data, intervention variants, model predictions, and evaluation code---so the same diagnostic can be applied to new VLMs.

\section*{Limitations}

\paragraph{What an intervention can and cannot say.} Our paired interventions identify \emph{which} channels a model uses but not exactly \emph{how} they are combined: e.g., the wrong\_object collapse for Qwen3-VL-235B shows that target identity is load-bearing, but does not isolate whether the underlying signal is shape, texture, or local context that co-varies with identity. Similarly, the apparent-size invariance under zoom-in is consistent with several mechanisms (e.g., category-conditioned readout that overrides apparent-size cues, or a vision encoder that resamples to a canonical scale before the language head sees it). Conclusions in the paper are stated at the level of cue use, not mechanism.

\paragraph{Closed-API reference.} ChatGPT~5.5 was queried through the provider API at a fixed point in time; we cannot guarantee that re-running these probes against the same endpoint months later will reproduce the exact numbers. For the open-weight models the inference is fully local and reproducible from the released code. We use the ChatGPT~5.5 row as a language-prior reference, not as a measurement of permanent capability.

\paragraph{Coverage gaps and language scope.} Molmo-7B-D refuses text-only and is reported on gray-blank; all other models are reported on text-only. All evaluations use English-language queries; we do not measure how the results transfer to other languages.

\section*{Ethical Considerations}

The ITW split consists of $70$ author-captured photographs of everyday objects with no identifiable people in frame; ground-truth dimensions were tape-measured. The Objectron split is used under its original license. Upon acceptance we will release Metric VQA (images, dimensions, intervention variants, predictions, evaluation code).




\bibliography{custom}

@article{quantiphy,
    title   = {{QuantiPhy}: A Quantitative Benchmark Evaluating Physical Reasoning Abilities of Vision-Language Models},
    author  = {Li, Puyin and Xiang, Tiange and Mao, Ella and Wei, Shirley and Chen, Xinye and Masood, Adnan and Li, Fei-Fei and Adeli, Ehsan},
    journal = {arXiv preprint arXiv:2512.19526},
    year    = {2025}
}

@article{depthlm,
    title   = {{DepthLM}: Metric Depth From Vision Language Models},
    author  = {Cai, Zhipeng and Yeh, Ching-Feng and Xu, Hu and Liu, Zhuang and Meyer, Gregory and Lei, Xinjie and Zhao, Changsheng and Li, Shang-Wen and Chandra, Vikas and Shi, Yangyang},
    journal = {arXiv preprint arXiv:2509.25413},
    year    = {2025}
}

@inproceedings{eigen2014depth,
    title     = {Depth Map Prediction from a Single Image using a Multi-Scale Deep Network},
    author    = {Eigen, David and Puhrsch, Christian and Fergus, Rob},
    booktitle = {Advances in Neural Information Processing Systems},
    year      = {2014}
}

@inproceedings{chen2024spatialvlm,
    title     = {{SpatialVLM}: Endowing Vision-Language Models with Spatial Reasoning Capabilities},
    author    = {Chen, Boyuan and Xu, Zhuo and Kirmani, Sean and Ichter, Brian and Sadigh, Dorsa and Guibas, Leonidas and Xia, Fei},
    booktitle = {Proceedings of the IEEE/CVF Conference on Computer Vision and Pattern Recognition (CVPR)},
    year      = {2024}
}

@inproceedings{liao2024qspatial,
    title     = {Reasoning Paths with Reference Objects Elicit Quantitative Spatial Reasoning in Large Vision-Language Models},
    author    = {Liao, Yuan-Hong and Mahmood, Rafid and Fidler, Sanja and Acuna, David},
    booktitle = {Proceedings of the 2024 Conference on Empirical Methods in Natural Language Processing (EMNLP)},
    year      = {2024}
}

@inproceedings{goyal2017vqa,
    title     = {Making the {V} in {VQA} Matter: Elevating the Role of Image Understanding in Visual Question Answering},
    author    = {Goyal, Yash and Khot, Tejas and Summers-Stay, Douglas and Batra, Dhruv and Parikh, Devi},
    booktitle = {Proceedings of the IEEE/CVF Conference on Computer Vision and Pattern Recognition (CVPR)},
    year      = {2017}
}

@inproceedings{agrawal2018gvqa,
    title     = {Don't Just Assume; Look and Answer: Overcoming Priors for Visual Question Answering},
    author    = {Agrawal, Aishwarya and Batra, Dhruv and Parikh, Devi and Kembhavi, Aniruddha},
    booktitle = {Proceedings of the IEEE/CVF Conference on Computer Vision and Pattern Recognition (CVPR)},
    year      = {2018}
}

@inproceedings{hsieh2023sugarcrepe,
    title     = {{SugarCrepe}: Fixing Hackable Benchmarks for Vision-Language Compositionality},
    author    = {Hsieh, Cheng-Yu and Zhang, Jieyu and Ma, Zixian and Kembhavi, Aniruddha and Krishna, Ranjay},
    booktitle = {Advances in Neural Information Processing Systems (Datasets and Benchmarks Track)},
    year      = {2023}
}

@article{zhao2022vlchecklist,
    title   = {{VL-CheckList}: Evaluating Pre-trained Vision-Language Models with Objects, Attributes and Relations},
    author  = {Zhao, Tiancheng and Zhang, Tianqi and Zhu, Mingwei and Shen, Haozhan and Lee, Kyusong and Lu, Xiaopeng and Yin, Jianwei},
    journal = {arXiv preprint arXiv:2207.00221},
    year    = {2022}
}

@inproceedings{fong2017perturbation,
    title     = {Interpretable Explanations of Black Boxes by Meaningful Perturbation},
    author    = {Fong, Ruth C. and Vedaldi, Andrea},
    booktitle = {Proceedings of the IEEE/CVF International Conference on Computer Vision (ICCV)},
    year      = {2017}
}

@inproceedings{ahmadyan2021objectron,
    title     = {Objectron: A Large Scale Dataset of Object-Centric Videos in the Wild With Pose Annotations},
    author    = {Ahmadyan, Adel and Zhang, Liangkai and Ablavatski, Artsiom and Wei, Jianing and Grundmann, Matthias},
    booktitle = {Proceedings of the IEEE/CVF Conference on Computer Vision and Pattern Recognition (CVPR)},
    year      = {2021},
    pages     = {7822--7831}
}

@inproceedings{masry2022chartqa,
    title     = {{ChartQA}: A Benchmark for Question Answering about Charts with Visual and Logical Reasoning},
    author    = {Masry, Ahmed and Long, Do Xuan and Tan, Jia Qing and Joty, Shafiq and Hoque, Enamul},
    booktitle = {Findings of the Association for Computational Linguistics: ACL 2022},
    pages     = {2263--2279},
    year      = {2022}
}

@misc{openai2024chatgpt,
    title        = {{ChatGPT}},
    author       = {{OpenAI}},
    year         = {2024},
    howpublished = {\url{https://chat.openai.com}}
}

@article{bai2025qwen25vl,
    title   = {Qwen2.5-VL Technical Report},
    author  = {Bai, Shuai and Chen, Keqin and Liu, Xuejing and Wang, Jialin and Ge, Wenbin and Song, Sibo and others},
    journal = {arXiv preprint arXiv:2502.13923},
    year    = {2025}
}

@inproceedings{chen2024internvl,
    title     = {{InternVL}: Scaling up Vision Foundation Models and Aligning for Generic Visual-Linguistic Tasks},
    author    = {Chen, Zhe and Wu, Jiannan and Wang, Wenhai and Su, Weijie and Chen, Guo and Xing, Sen and others},
    booktitle = {Proceedings of the IEEE/CVF Conference on Computer Vision and Pattern Recognition (CVPR)},
    year      = {2024}
}

@inproceedings{liu2023llava,
    title     = {Visual Instruction Tuning},
    author    = {Liu, Haotian and Li, Chunyuan and Wu, Qingyang and Lee, Yong Jae},
    booktitle = {Advances in Neural Information Processing Systems},
    year      = {2023}
}

@inproceedings{deitke2025molmo,
    title     = {{Molmo} and {PixMo}: Open Weights and Open Data for State-of-the-Art Vision-Language Models},
    author    = {Deitke, Matt and Clark, Christopher and Lee, Sangho and others},
    booktitle = {Proceedings of the IEEE/CVF Conference on Computer Vision and Pattern Recognition (CVPR)},
    year      = {2025}
}

@article{marafioti2025smolvlm,
    title   = {{SmolVLM}: Redefining Small and Efficient Multimodal Models},
    author  = {Marafioti, Andr{\'e}s and Zohar, Orr and Farr{\'e}, Miquel and others},
    journal = {arXiv preprint arXiv:2504.05299},
    year    = {2025}
}

@inproceedings{hu2021lora,
    title     = {{LoRA}: Low-Rank Adaptation of Large Language Models},
    author    = {Hu, Edward J. and Shen, Yelong and Wallis, Phillip and Allen-Zhu, Zeyuan and Li, Yuanzhi and Wang, Shean and Wang, Lu and Chen, Weizhu},
    booktitle = {International Conference on Learning Representations (ICLR)},
    year      = {2022}
}

@article{carion2025sam3,
    title     = {{SAM} 3: Segment Anything with Concepts},
    author    = {Carion, Nicolas and Gustafson, Laura and Hu, Yuan-Ting and Debnath, Shoubhik and Hu, Ronghang and Suris, Didac and Ryali, Chaitanya and Alwala, Kalyan Vasudev and Khedr, Haitham and Huang, Andrew and Lei, Jie and Ma, Tengyu and Guo, Baishan and Kalla, Arpit and Marks, Markus and Greer, Joseph and Wang, Meng and Sun, Peize and R{\"a}dle, Roman and Afouras, Triantafyllos and Mavroudi, Effrosyni and Xu, Katherine and Wu, Tsung-Han and Zhou, Yu and Momeni, Liliane and Hazra, Rishi and Ding, Shuangrui and Vaze, Sagar and Porcher, Francois and Li, Feng and Li, Siyuan and Kamath, Aishwarya and Cheng, Ho Kei and Doll{\'a}r, Piotr and Ravi, Nikhila and Saenko, Kate and Zhang, Pengchuan and Feichtenhofer, Christoph},
    year      = {2025},
    journal   = {arXiv preprint arXiv:2511.16719}
}

\appendix

\section{Additional Results}
\label{sec:appendix}

\begin{figure*}[t]
\centering
\includegraphics[width=\textwidth]{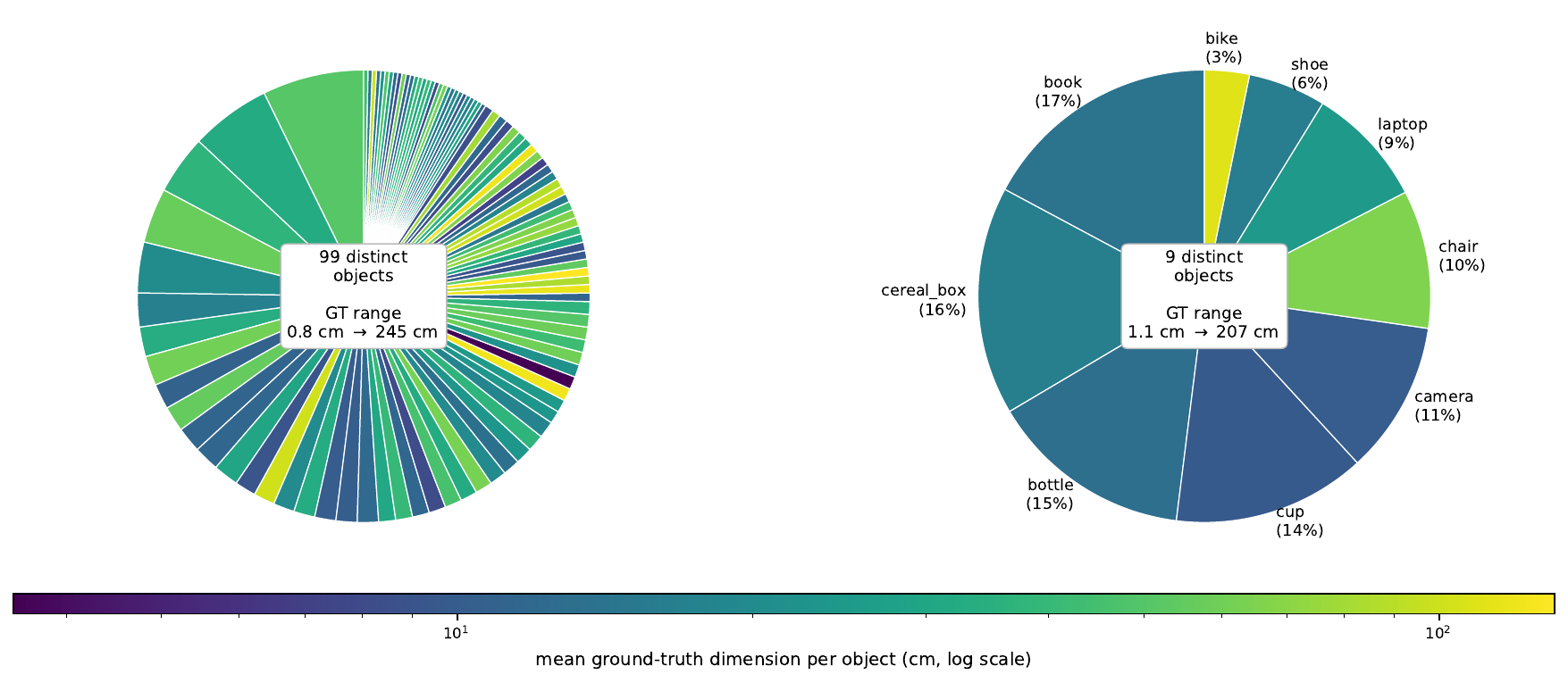}
\caption{\textbf{Object-class composition of the two splits.} Each slice is one distinct object class, sized by query count and colored by mean GT on a shared log-cm scale. Left: in-the-wild ($n{=}331$) contains $99$ distinct objects. Right: Objectron ($n{=}10{,}482$) contains $9$.}
\label{fig:dataset_composition}
\end{figure*}

\begin{figure}[t]
\centering
\includegraphics[width=\columnwidth]{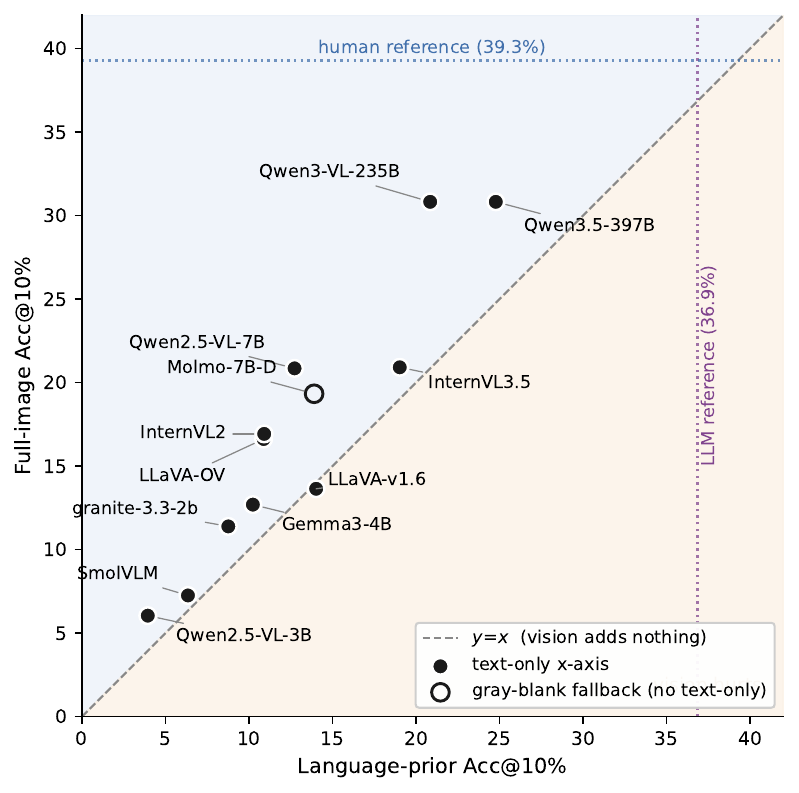}
\caption{\textbf{Image-conditioned versus language-prior Acc@10\% on the in-the-wild split.} $x$: each model's language-prior Acc (text-only, or gray-blank fallback). $y$: same model's Acc with the original image. Points above the diagonal: image helped. Dotted purple line: ChatGPT~5.5 text-only.}
\label{fig:prior_vs_pixel}
\end{figure}

\begin{figure*}[t]
\centering
\includegraphics[width=\textwidth]{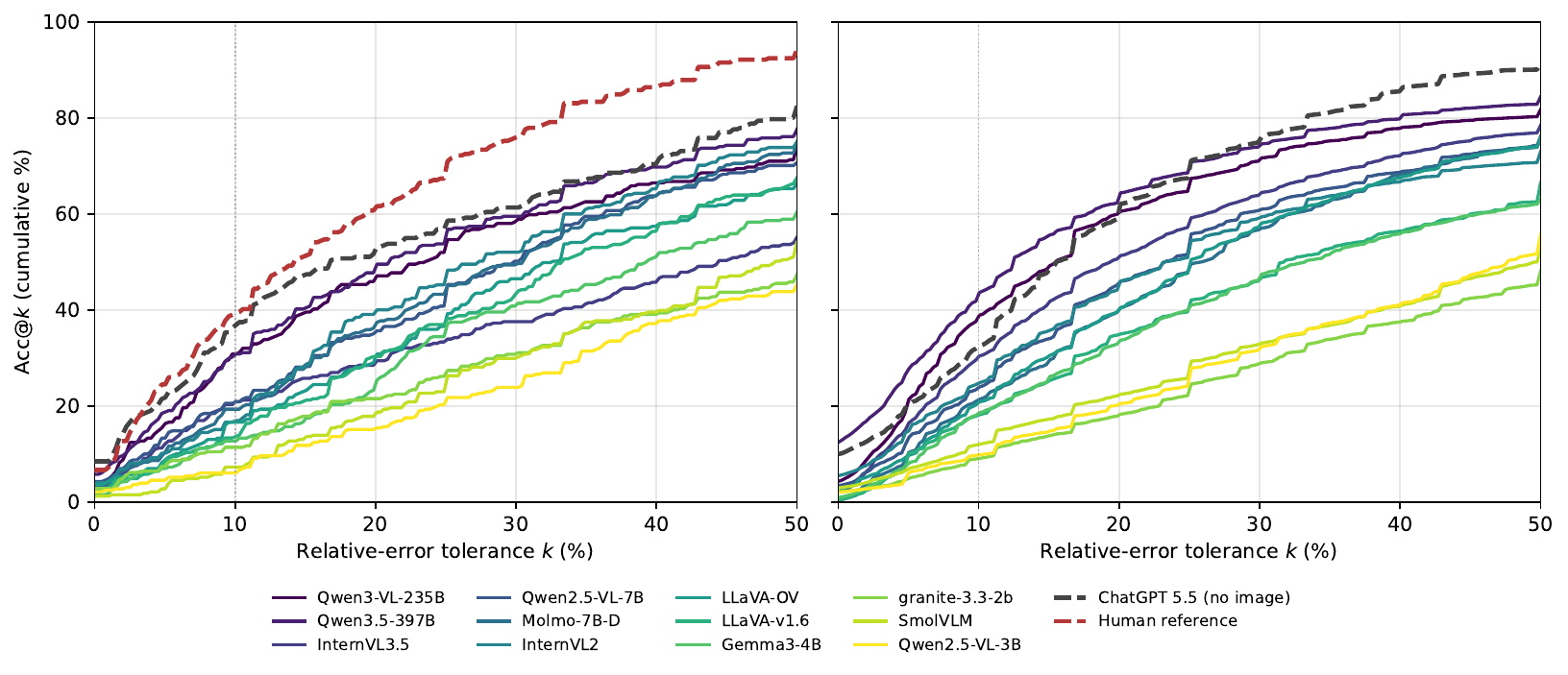}
\caption{\textbf{Cumulative accuracy as a function of the relative-error tolerance $k$.} $x$: tolerance $k \in [0, 50]\%$. $y$: fraction of queries with $|\hat{y}-y|/|y| \leq k$. Dotted vertical at $k{=}10\%$ is the headline metric. Dashed lines: ChatGPT~5.5 text-only (both splits); human reference (in-the-wild only).}
\label{fig:acc_at_k}
\end{figure*}

This appendix contains per-model breakdowns and supporting visualizations referenced by the body. \emph{Dataset composition:} Figure~\ref{fig:dataset_composition} contrasts the object-class composition and physical-size variance of the two splits. \emph{MAE counterpart:} Table~\ref{tab:main_mae} reports the MAE version of Table~\ref{tab:main} for both splits. \emph{Language priors:} Table~\ref{tab:language_prior} reports per-model vision / gray-blank / text-only Acc@10\% on both splits. Molmo-7B-D refuses text-only and LLaVA-v1.6 refuses gray-blank, accounting for the "--" entries. Figure~\ref{fig:prior_vs_pixel} plots image-conditioned Acc against language-prior Acc on ITW split. \emph{Failure-mode decomposition:} Table~\ref{tab:fse_itw} decomposes each model's ITW predictions into parse failures, scale failures, and committed estimations, and reports Acc@10\% conditional on the committed subset. \emph{Cumulative accuracy:} Figure~\ref{fig:acc_at_k} sweeps the tolerance $k$ from $0$ to $50$\% to show that the qualitative ordering in Table~\ref{tab:main} is not specific to $k{=}10\%$. \emph{Context-variant decomposition:} Table~\ref{tab:context} reports per-model Acc@10\% on the four mask-fill canvases (object\_only, object\_ring, wrong\_object, random\_mask); Table~\ref{tab:context_deltas_ci} reports paired-bootstrap $95$\% CIs on the three load-bearing context deltas referenced in Section~\ref{sec:results}---wrong\_object collapse, ring lift, and target-pixel content (object\_only $-$ random\_mask). \emph{Apparent size:} Tables~\ref{tab:zoom_in} and~\ref{tab:zoom_out} report each model's MAE and Acc@10\% under the zoom-in $1.4\times$ and zoom-out $0.6\times$ object-only variants, both compared against the per-model text-only and \textit{object\_only} anchors. Table~\ref{tab:apparent_size_bias} reports the compact paired-ratio analysis referenced in Section~\ref{sec:results}. \emph{Lens distortion:} Table~\ref{tab:distortion} reports the per-model paired $\Delta$\,Acc@10\% under both $|k|=0.15$ distortions discussed in Section~\ref{sec:results}.

\begin{table}[t]
\centering\small
\begin{tabular}{l|cc}
\toprule
Model & ITW MAE (cm) & Obj MAE (cm) \\
 & ($n{=}331$) & ($n{=}10{,}482$) \\
\midrule
\multicolumn{3}{l}{\emph{Non-VLM reference baselines}} \\
Human reference & 6.48 & -- \\
ChatGPT 5.5 (text) & 9.53 & 4.06 \\
\midrule
Qwen3-VL-235B & 10.88 & 5.02 \\
Qwen3.5-397B & 10.27 & 4.38 \\
InternVL3.5-241B & 17.42 & 5.72 \\
Qwen2.5-VL-7B & 30.60 & 7.34 \\
Molmo-7B-D & 13.31 & 9.74 \\
InternVL2 & 19.10 & 23.39 \\
LLaVA-OV & 24.31 & 10.25 \\
LLaVA-v1.6 & 16.86 & 9.18 \\
Gemma3-4B & 19.91 & 12.03 \\
granite-3.3-2b & 46.40 & 50.08 \\
SmolVLM & 28.38 & 49.16 \\
Qwen2.5-VL-3B & 25.30 & 12.19 \\
\bottomrule
\end{tabular}
\caption{Mean absolute error (cm) counterpart to Table~\ref{tab:main}, rows in the same order (in-the-wild Acc@10\% descending). MAE on in-the-wild is dominated by outliers over the $0.8$--$245$\,cm ground-truth range and does not match the Acc@10\% ordering (Spearman $\rho = 0.25$, $p > 0.4$); we therefore use Acc@10\% as the primary metric in the main text.}
\label{tab:main_mae}
\end{table}

\begin{table*}[t]
\centering\small
\begin{tabular}{l|ccc|ccc}
\toprule
 & \multicolumn{3}{c|}{In-the-wild Acc@10\%} & \multicolumn{3}{c}{Objectron Acc@10\%} \\
Model & vision & gray-blank & text-only & vision & gray-blank & text-only \\
\midrule
ChatGPT 5.5 LLM (no image) & -- & -- & 36.9\% & -- & -- & 32.4\% \\
\midrule
Qwen2.5-VL-3B & 6.0\% & 4.5\% & 3.9\% & 9.8\% & 9.8\% & 4.7\% \\
LLaVA-v1.6 & 13.6\% & -- & 14.0\% & 18.1\% & -- & 19.3\% \\
SmolVLM & 7.3\% & 6.9\% & 6.3\% & 12.0\% & 7.0\% & 11.5\% \\
Gemma3-4B & 12.7\% & 9.1\% & 10.2\% & 18.5\% & 15.0\% & 21.7\% \\
granite-3.3-2b & 11.4\% & 7.6\% & 8.8\% & 9.0\% & 11.7\% & 16.7\% \\
Qwen2.5-VL-7B & 20.8\% & 5.2\% & 12.7\% & 23.9\% & 15.7\% & 22.0\% \\
LLaVA-OV & 16.6\% & 9.1\% & 10.9\% & 20.8\% & 14.9\% & 17.0\% \\
Molmo-7B-D & 19.3\% & 13.9\% & -- & 21.4\% & 28.1\% & -- \\
InternVL2 & 16.9\% & 6.4\% & 10.9\% & 24.8\% & 18.5\% & 20.4\% \\
InternVL3.5-241B & 20.9\% & 15.8\% & 19.0\% & 30.3\% & 21.6\% & 24.6\% \\
Qwen3-VL-235B & 30.8\% & 10.3\% & 20.8\% & 38.9\% & 2.8\% & 22.5\% \\
Qwen3.5-397B & 30.8\% & 4.8\% & 24.8\% & 43.6\% & 1.2\% & 30.3\% \\
\bottomrule
\end{tabular}
\caption{Language-only prior probe on both splits. The conceptually clean intervention is \emph{text-only} (no image in the prompt at all). Molmo-7B-D refuses text-only entirely (returns empty on every query, \textit{--} in the text-only columns); LLaVA-v1.6-Mistral-7B refuses gray-blank instead, producing parseable numeric output on only one of $331$ in-the-wild queries. \emph{Gray-blank} (a uniform gray canvas at RGB 127) supplies the image slot the API requires while providing no visual evidence about the target. For the smallest open VLMs gray-blank and text-only agree within a few percentage points, but for the two largest open models supplying a gray canvas suppresses the LM prior far more than removing the image entirely (e.g., Qwen3.5-397B on Objectron: $1.2\%$ gray vs $30.3\%$ text-only). We therefore read text-only as the primary anchor where available and gray-blank as a conservative fallback otherwise. Neither probe closes the gap with the frontier text-only LLM (ChatGPT~5.5), indicating that open VLMs' internal language priors are largely inaccessible under the VLM API.}
\label{tab:language_prior}
\end{table*}

\paragraph{Decomposing predictions into failure modes.} The headline Acc@10\% numbers conflate two qualitatively different failure modes: parse failures, in which the model produces no numeric answer at all, and scale failures, in which the model commits to a number off by more than the true value. Table~\ref{tab:fse_itw} decomposes each model's predictions into \emph{F} (parse failure), \emph{S} (scale failure, $|\hat{y}-y| > y$), and \emph{E} (estimation, $|\hat{y}-y| \leq y$), reporting MAE and Acc@10\% conditional on the $E$ subset (the predictions where the model committed to a plausibly-scaled answer). Three observations: (i) All twelve models engage on essentially every ITW query; the highest parse-failure rate in the pool is below $2$\%. (ii) Scale failures vary substantially, from $5.1$\% (InternVL2) up to $23.6$\% (granite-3.3-2b); a high $S$ rate signals a model that confidently produces wildly wrong numbers, distinct from refusal. (iii) Conditional on the $E$ subset, the two largest models lead the pool (Qwen3.5-397B at $33.3$\%, Qwen3-VL-235B at $33.2$\% Acc@10$_E$); the small-open pool tops out at Qwen2.5-VL-7B ($23.7$\%). The ranking by Acc@10$_E$ approximately matches the unconditional ranking but tightens the spread, suggesting that some of the unconditional Acc@10\% gap between models is driven by how often they commit at all rather than by how accurately they commit.

\begin{table*}[t]
\centering\small
\begin{tabular}{l|r|rrr|r|rr}
\toprule
Model & $N$ & \%F & \%S & \%E & med.\,AE$_{S}$ & MAE$_{E}$ & Acc@10$_{E}$ \\
\midrule
Qwen3.5-397B & 331 & 0.0 & 7.6 & 92.4 & 24.70 & 8.41 & 33.3\% \\
Qwen3-VL-235B & 331 & 0.0 & 7.3 & 92.7 & 32.80 & 8.59 & 33.2\% \\
Qwen2.5-VL-7B & 331 & 0.0 & 12.1 & 87.9 & 91.00 & 10.85 & 23.7\% \\
LLaVA-OV & 331 & 0.0 & 19.6 & 80.4 & 80.00 & 11.80 & 20.7\% \\
Molmo-7B-D & 331 & 1.5 & 8.5 & 90.0 & 10.47 & 11.89 & 21.1\% \\
InternVL3.5-241B & 331 & 0.3 & 21.1 & 78.5 & 28.50 & 12.19 & 26.5\% \\
Gemma3-4B & 331 & 0.0 & 19.9 & 80.1 & 36.50 & 13.52 & 15.8\% \\
InternVL2 & 331 & 1.8 & 5.1 & 93.1 & 14.50 & 15.15 & 17.9\% \\
LLaVA-v1.6 & 331 & 0.3 & 3.6 & 96.1 & 55.75 & 15.80 & 14.2\% \\
SmolVLM & 331 & 0.0 & 9.1 & 90.9 & 62.25 & 21.35 & 8.0\% \\
granite-3.3-2b & 331 & 0.3 & 23.6 & 76.1 & 77.25 & 21.48 & 14.7\% \\
Qwen2.5-VL-3B & 331 & 0.0 & 8.8 & 91.2 & 25.50 & 22.20 & 6.6\% \\
\bottomrule
\end{tabular}
\caption{\textbf{Failure / Scale / Estimation (F/S/E) decomposition on the in-the-wild split.} Each model's $N$ predictions partition into \textbf{F} = parse failure (no numeric in the response, including refusals like ``cannot measure without a reference''); \textbf{S} = scale failure ($|\hat{y} - y| > y$; the model commits to a number off by more than the true value); and \textbf{E} = estimation ($|\hat{y} - y| \leq y$). The last three columns describe the $E$ subset: median absolute error within $S$ (for context on how bad the wrong numbers are), MAE on $E$ (typical accuracy \emph{conditional on the model committing to a reasonable answer}), and the corresponding Acc@10\%. Open VLMs are sorted by ascending MAE$_E$. The decomposition separates ``the model doesn't engage'' from ``the model engages but is miscalibrated'': both contribute to a model's headline MAE differently and warrant different mitigations.}
\label{tab:fse_itw}
\end{table*}

\begin{table*}[t]
\centering\small
\begin{tabular}{l|c|ccccc}
\toprule
Model & text-only & full & object\_only & object\_ring & wrong\_object & random\_mask \\
\midrule
Qwen2.5-VL-3B & 3.9\% & 6.4\% & 8.2\% & 8.8\% & 3.2\% & 5.0\% \\
LLaVA-v1.6 & 14.0\% & 12.7\% & 3.3\% & 7.2\% & 0.9\% & 1.5\% \\
SmolVLM & 6.3\% & 7.3\% & 3.8\% & 7.6\% & 2.9\% & 3.8\% \\
Gemma3-4B & 10.2\% & 12.3\% & 14.6\% & 18.1\% & 9.2\% & 9.7\% \\
granite-3.3-2b & 8.8\% & 11.3\% & 4.7\% & 9.5\% & 2.3\% & 5.2\% \\
Qwen2.5-VL-7B & 12.7\% & 21.1\% & 14.2\% & 18.1\% & 3.6\% & 5.1\% \\
LLaVA-OV & 10.9\% & 16.1\% & 10.6\% & 17.3\% & 4.8\% & 10.6\% \\
Molmo-7B-D & -- & 19.4\% & 12.4\% & 16.4\% & 10.5\% & 11.7\% \\
InternVL2 & 10.9\% & 17.0\% & 11.6\% & 12.9\% & 6.8\% & 9.1\% \\
InternVL3.5-241B & 19.0\% & 19.6\% & 25.2\% & 22.4\% & 14.5\% & 18.6\% \\
Qwen3-VL-235B & 20.8\% & 30.7\% & 33.0\% & 31.6\% & 11.1\% & 17.0\% \\
Qwen3.5-397B & 24.8\% & 30.4\% & 23.7\% & 25.1\% & 9.6\% & 13.7\% \\
\bottomrule
\end{tabular}
\caption{Context-variant probe on in-the-wild (Acc@10\%). The \textit{text-only} column is the per-model LM-prior anchor (no image sent to the model at all). The five context variants share the \textit{object\_only} canvas (gray fill outside the SAM mask) but differ in what visual evidence is added: \emph{full} = full scene; \emph{object\_only} = target mask only; \emph{object\_ring} = target + 50px halo; \emph{wrong\_object} = a different object's mask swapped in; \emph{random\_mask} = a random rectangle of matching area. Reading each variant against text-only tells us what the added visual evidence buys: target identity (\textit{object\_only} vs text-only), local context (\textit{object\_ring} vs \textit{object\_only}), and the cost of mis-identifying the target (\textit{wrong\_object} vs \textit{object\_only}).}
\label{tab:context}
\end{table*}

\begin{table*}[t]
\centering\small
\begin{tabular}{l|ccc}
\toprule
Model & $\Delta$ wrong\_obj. & $\Delta$ ring & $\Delta$ target pixels \\
 & (vs object\_only) & (vs object\_only) & (object\_only $-$ random\_mask) \\
\midrule
Qwen2.5-VL-3B & $-5.2$\,{\scriptsize[-8.0,\,-2.7]} & $+0.3$\,{\scriptsize[-2.5,\,+3.3]} & $+3.3$\,{\scriptsize[+0.5,\,+6.0]} \\
LLaVA-v1.6 & $-1.8$\,{\scriptsize[-3.6,\,+0.0]} & $+4.6$\,{\scriptsize[+1.7,\,+7.7]} & $+1.4$\,{\scriptsize[-0.6,\,+3.5]} \\
SmolVLM & $-0.8$\,{\scriptsize[-3.0,\,+1.4]} & $+4.1$\,{\scriptsize[+1.1,\,+7.2]} & $+0.3$\,{\scriptsize[-1.9,\,+2.7]} \\
Gemma3-4B & $-5.4$\,{\scriptsize[-9.5,\,-1.2]} & $+3.8$\,{\scriptsize[-0.3,\,+8.0]} & $+4.7$\,{\scriptsize[+0.3,\,+9.2]} \\
granite-3.3-2b & $-1.8$\,{\scriptsize[-4.3,\,+0.3]} & $+4.8$\,{\scriptsize[+2.3,\,+7.6]} & $-0.6$\,{\scriptsize[-3.2,\,+2.0]} \\
Qwen2.5-VL-7B & $-10.6$\,{\scriptsize[-14.2,\,-6.9]} & $+3.9$\,{\scriptsize[+0.0,\,+8.2]} & $+9.1$\,{\scriptsize[+5.4,\,+13.0]} \\
LLaVA-OV & $-5.8$\,{\scriptsize[-9.4,\,-2.4]} & $+6.7$\,{\scriptsize[+3.0,\,+10.3]} & $+0.0$\,{\scriptsize[-3.6,\,+3.9]} \\
Molmo-7B-D & $-1.7$\,{\scriptsize[-5.7,\,+2.4]} & $+4.0$\,{\scriptsize[+0.6,\,+7.3]} & $+0.9$\,{\scriptsize[-2.8,\,+4.3]} \\
InternVL2 & $-4.5$\,{\scriptsize[-7.9,\,-1.0]} & $+1.2$\,{\scriptsize[-1.5,\,+4.3]} & $+2.5$\,{\scriptsize[-0.9,\,+6.0]} \\
InternVL3.5-241B & $-9.6$\,{\scriptsize[-15.4,\,-3.9]} & $-2.6$\,{\scriptsize[-7.1,\,+1.8]} & $+6.3$\,{\scriptsize[+1.3,\,+11.2]} \\
Qwen3-VL-235B & $-20.3$\,{\scriptsize[-25.5,\,-15.1]} & $-1.1$\,{\scriptsize[-4.9,\,+2.7]} & $+15.7$\,{\scriptsize[+11.0,\,+20.1]} \\
Qwen3.5-397B & $-12.4$\,{\scriptsize[-17.3,\,-8.0]} & $+2.7$\,{\scriptsize[-1.9,\,+7.4]} & $+9.3$\,{\scriptsize[+4.7,\,+13.7]} \\
\bottomrule
\end{tabular}
\caption{\textbf{Paired-bootstrap CIs on the three load-bearing context deltas} (in-the-wild). Each cell is the per-model Acc@10\% $\Delta$ between two context variants on the paired intersection of (query, answer) rows, with a $95$\% paired bootstrap CI ($B{=}2000$ resamples; MD5-stable seed). \textbf{wrong\_obj.}: replacing the target's mask with a different object's collapses Acc@10\% (CI strictly below $0$) for the models that use target identity. \textbf{ring}: adding a $50$\,px halo lifts Acc@10\% (CI strictly above $0$) for the small-to-mid pool. \textbf{target pixels}: the gap object\_only $-$ random\_mask is positive when the target's actual pixel content carries information beyond its area; the CI distinguishes models that genuinely use target pixels from those tied with a matched-area random patch.}
\label{tab:context_deltas_ci}
\end{table*}

\begin{table*}[t]
\centering\small
\begin{tabular}{l|c|c|c}
\toprule
Model & med ratio & \% rows ratio $<$ 1 & direction \\
 & (zoom-out / obj\_only) & & \\
\midrule
Qwen2.5-VL-3B & 1.00 & 35.6\% & invariant \\
LLaVA-v1.6 & 0.80 & 54.2\% & invariant \\
SmolVLM & 1.00 & 33.4\% & invariant \\
Gemma3-4B & 0.80 & 64.1\% & shrinks $\downarrow$ \\
granite-3.3-2b & 1.00 & 36.0\% & invariant \\
Qwen2.5-VL-7B & 1.00 & 41.8\% & invariant \\
LLaVA-OV & 0.83 & 59.9\% & shrinks $\downarrow$ \\
Molmo-7B-D & 1.00 & 36.2\% & invariant \\
InternVL2 & 1.00 & 43.0\% & invariant \\
InternVL3.5-241B & 0.95 & 61.3\% & invariant \\
Qwen3-VL-235B & 1.00 & 24.7\% & invariant \\
Qwen3.5-397B & 1.00 & 32.2\% & invariant \\
\bottomrule
\end{tabular}
\caption{Apparent-size bias: zoom-out $0.6\times$ vs \textit{object\_only} (same image, same identity, only apparent pixel size differs). We report the scale-free median paired ratio $r = \mathrm{med}(\hat{y}_{\mathrm{zoom\text{-}out}} / \hat{y}_{\mathrm{object\_only}})$ across paired rows. $r = 1$ is invariance; $r < 1$ means the model shrinks its prediction when the target is rendered smaller. The \textit{direction} column codes the per-model verdict: \textbf{shrinks $\downarrow$} when $r \leq 0.90$ (at least 10\% shrinkage) and $\geq 55$\% of paired rows decrease; \textbf{grows $\uparrow$} for the symmetric case; \textbf{invariant} otherwise.}
\label{tab:apparent_size_bias}
\end{table*}

\begin{table*}[t]
\centering\small
\begin{tabular}{l|cc|cc}
\toprule
Model & text-only & object\_only & \multicolumn{2}{c}{zoom-in 1.4$\times$} \\
 & Acc@10 & Acc@10 & MAE & Acc@10 \\
\midrule
Qwen2.5-VL-3B & 3.9\% & 8.2\% & 21.94 & 8.8\% \\
LLaVA-v1.6 & 14.0\% & 3.3\% & 21.18 & 6.8\% \\
SmolVLM & 6.3\% & 3.8\% & 40.23 & 3.8\% \\
Gemma3-4B & 10.2\% & 14.6\% & 15.00 & 13.3\% \\
granite-3.3-2b & 8.8\% & 4.7\% & 31.68 & 5.4\% \\
Qwen2.5-VL-7B & 12.7\% & 14.2\% & 17.38 & 12.6\% \\
LLaVA-OV & 10.9\% & 10.6\% & 20.97 & 11.6\% \\
Molmo-7B-D & -- & 12.4\% & 15.93 & 14.2\% \\
InternVL2 & 10.9\% & 11.6\% & 20.18 & 12.2\% \\
InternVL3.5-241B & 19.0\% & 25.2\% & 13.17 & 23.2\% \\
Qwen3-VL-235B & 20.8\% & 33.0\% & 12.08 & 30.8\% \\
Qwen3.5-397B & 24.8\% & 23.7\% & 15.35 & 23.9\% \\
\bottomrule
\end{tabular}
\caption{Zoom-in $1.4\times$ on object-only images. The target object is scaled up by 40\% on the gray canvas; no scene or reference cues. Compared against the \textit{object\_only} anchor (same image, no zoom). Most VLMs are invariant to this magnification, consistent with category-prior-based estimation rather than pixel-area readout.}
\label{tab:zoom_in}
\end{table*}

\begin{table*}[t]
\centering\small
\begin{tabular}{l|cc|cc}
\toprule
Model & text-only & object\_only & \multicolumn{2}{c}{zoom-out 0.6$\times$} \\
 & Acc@10 & Acc@10 & MAE & Acc@10 \\
\midrule
Qwen2.5-VL-3B & 3.9\% & 8.2\% & 23.01 & 7.5\% \\
LLaVA-v1.6 & 14.0\% & 3.3\% & 26.91 & 2.6\% \\
SmolVLM & 6.3\% & 3.8\% & 25.70 & 3.5\% \\
Gemma3-4B & 10.2\% & 14.6\% & 17.57 & 8.5\% \\
granite-3.3-2b & 8.8\% & 4.7\% & 26.97 & 3.8\% \\
Qwen2.5-VL-7B & 12.7\% & 14.2\% & 18.33 & 9.4\% \\
LLaVA-OV & 10.9\% & 10.6\% & 21.90 & 7.2\% \\
Molmo-7B-D & -- & 12.4\% & 17.15 & 13.3\% \\
InternVL2 & 10.9\% & 11.6\% & 26.01 & 9.0\% \\
InternVL3.5-241B & 19.0\% & 25.2\% & 12.24 & 21.4\% \\
Qwen3-VL-235B & 20.8\% & 33.0\% & 11.25 & 31.1\% \\
Qwen3.5-397B & 24.8\% & 23.7\% & 16.61 & 22.3\% \\
\bottomrule
\end{tabular}
\caption{Zoom-out $0.6\times$ on object-only images. The target object is scaled down by 40\%; remaining canvas filled gray. Compared against the \textit{object\_only} anchor. See Table~\ref{tab:apparent_size_bias} for the paired in/out ratio test of apparent-size bias.}
\label{tab:zoom_out}
\end{table*}

\begin{table*}[t]
\centering\small
\begin{tabular}{l|c|cc|c}
\toprule
Model & full Acc@10\% & $\Delta$ barrel $k{=}0.15$ & $\Delta$ pincushion $k{=}-0.15$ & verdict \\
\midrule
Qwen2.5-VL-3B & 6.0\% & $+0.9$\,{\scriptsize[-2.2,\,+4.1]} & $+7.5$\,{\scriptsize[+3.8,\,+11.3]} & mixed \\
LLaVA-v1.6 & 13.7\% & $-2.5$\,{\scriptsize[-5.4,\,+0.0]} & $-1.3$\,{\scriptsize[-4.1,\,+1.6]} & invariant \\
SmolVLM & 7.5\% & $+1.6$\,{\scriptsize[-0.6,\,+3.8]} & $+2.5$\,{\scriptsize[+0.0,\,+5.0]} & invariant \\
Gemma3-4B & 12.9\% & $+4.4$\,{\scriptsize[+0.6,\,+8.5]} & $+1.6$\,{\scriptsize[-2.5,\,+5.7]} & mixed \\
granite-3.3-2b & 12.2\% & $-2.2$\,{\scriptsize[-5.4,\,+0.6]} & $-2.6$\,{\scriptsize[-6.1,\,+0.6]} & invariant \\
Qwen2.5-VL-7B & 21.4\% & $-3.5$\,{\scriptsize[-6.6,\,-0.3]} & $-5.3$\,{\scriptsize[-9.1,\,-1.6]} & all hurt $\downarrow$ \\
LLaVA-OV & 17.9\% & $-0.9$\,{\scriptsize[-4.4,\,+2.5]} & $-0.9$\,{\scriptsize[-4.1,\,+2.5]} & invariant \\
Molmo-7B-D & 19.8\% & $+1.3$\,{\scriptsize[-1.9,\,+4.2]} & $-1.6$\,{\scriptsize[-5.1,\,+1.9]} & invariant \\
InternVL2 & 17.0\% & $+0.3$\,{\scriptsize[-2.2,\,+2.9]} & $+1.3$\,{\scriptsize[-1.9,\,+4.8]} & invariant \\
InternVL3.5-241B & 21.2\% & $-1.2$\,{\scriptsize[-4.5,\,+1.8]} & $-0.9$\,{\scriptsize[-4.5,\,+2.7]} & invariant \\
Qwen3-VL-235B & 31.0\% & $-0.6$\,{\scriptsize[-3.8,\,+2.6]} & $-2.9$\,{\scriptsize[-6.1,\,+0.3]} & invariant \\
Qwen3.5-397B & 30.7\% & $+3.2$\,{\scriptsize[-0.3,\,+6.7]} & $-5.0$\,{\scriptsize[-9.1,\,-0.9]} & mixed \\
\bottomrule
\end{tabular}
\caption{\textbf{Radial distortion probe (paired Acc@10\%).} Each image is warped with $r' = r(1 + k r^2)$ at $|k| = 0.15$ in both barrel and pincushion directions, with edge-preserving pre-shrink by $1/(1+|k|)$. $\Delta$ columns are Acc@10\% on the distorted condition minus Acc@10\% on the full image, computed on the paired intersection of (query, answer) rows present in both runs; the bracketed range is a $95$\% paired-bootstrap CI on $\Delta$ ($B{=}2000$ resamples of paired row indices). \textbf{verdict} classifies each model: \emph{invariant} when both CIs cover $0$; \emph{all hurt}/\emph{all help} when both CIs lie strictly below/above $0$; \emph{mixed} otherwise. Only one model (Qwen2.5-VL-7B) shows significant degradation under both directions (the closest signature of partial perspective-cue use); three models are \emph{mixed} (significant shift in one direction, CI covering zero in the other); the remaining eight are CI-invariant. The sign of the shift does not pattern with the direction of distortion (barrel vs.\ pincushion) for any model, ruling out vanishing-point-specific reasoning.}
\label{tab:distortion}
\end{table*}

\end{document}